\documentclass[sigconf]{acmart}

\usepackage{multirow}
\usepackage{hyperref}
\usepackage{siunitx}
\usepackage{float}
\usepackage{subcaption}
\AtBeginDocument{%
  \providecommand\BibTeX{{%
    \normalfont B\kern-0.5em{\scshape i\kern-0.25em b}\kern-0.8em\TeX}}}

\copyrightyear{2023} 
\acmYear{2023} 
\setcopyright{acmlicensed}\acmConference[ICAIF '23]{4th ACM International Conference on AI in Finance}{November 27--29, 2023}{Brooklyn, NY, USA}
\acmBooktitle{4th ACM International Conference on AI in Finance (ICAIF '23), November 27--29, 2023, Brooklyn, NY, USA}
\acmPrice{15.00}
\acmDOI{10.1145/3604237.3626905}
\acmISBN{979-8-4007-0240-2/23/11}

\begin{document}

\title{From Pixels to Predictions: Spectrogram and Vision Transformer for Better Time Series Forecasting}

\author{Zhen Zeng}
\affiliation{%
  \institution{J.~P.~Morgan AI Research}
  \city{New York}
  \state{NY}
  \country{USA}}
\email{zhen.zeng@jpmchase.com}

\author{Rachneet Kaur}
\authornote{Both authors contributed equally to this research.}
\affiliation{%
  \institution{J.~P.~Morgan AI Research}
  \city{New York}
  \state{NY}
  \country{USA}}
\email{rachneet.kaur@jpmorgan.com}

\author{Suchetha Siddagangappa}
\authornotemark[1]
\affiliation{%
  \institution{J.~P.~Morgan AI Research}
  \city{New York}
  \state{NY}
  \country{USA}}
\email{suchetha.siddagangappa@jpmchase.com}

\author{Tucker Balch}
\affiliation{%
  \institution{J.~P.~Morgan AI Research}
  \city{New York}
  \state{NY}
  \country{USA}}
\email{tucker.balch@jpmchase.com}

\author{Manuela Veloso}
\affiliation{%
  \institution{J.~P.~Morgan AI Research}
  \city{New York}
  \state{NY}
  \country{USA}}
\email{manuela.veloso@jpmchase.com}
\renewcommand{\shortauthors}{Zeng, et al.}







\begin{abstract}
Time series forecasting plays a crucial role in decision-making across various domains, but it presents significant challenges. Recent studies have explored image-driven approaches using computer vision models to address these challenges, often employing lineplots as the visual representation of time series data. In this paper, we propose a novel approach that uses time-frequency spectrograms as the visual representation of time series data. We introduce the use of a vision transformer for multimodal learning, showcasing the advantages of our approach across diverse datasets from different domains. To evaluate its effectiveness, we compare our method against statistical baselines (EMA and ARIMA), a state-of-the-art deep learning-based approach (DeepAR), other visual representations of time series data (lineplot images), and an ablation study on using only the time series as input. Our experiments demonstrate the benefits of utilizing spectrograms as a visual representation for time series data, along with the advantages of employing a vision transformer for simultaneous learning in both the time and frequency domains.
\end{abstract}

\maketitle
\section{Introduction}\label{sec:introduction}
Time series forecasting poses significant challenges due to the inherent noise in the data. Traditional statistical methods often utilize linear regression, exponential smoothing~\cite{holt2004forecasting, winters1960forecasting, gardner1985forecasting}, and autoregression models~\cite{makridakis2020m4}. In recent years, deep learning has witnessed substantial progress, focusing on ensemble models and sequence-to-sequence modeling, such as recurrent neural networks (RNNs), long short-term memory (LSTM)~\cite{hochreiter1997long}, and more recently, transformers~\cite{vaswani2017attention}. These advanced deep learning techniques offer enhanced capabilities in capturing complex non-linear dependencies within time series data.

\begin{figure}[t!]
    \centering
    \includegraphics[width=0.4\textwidth]{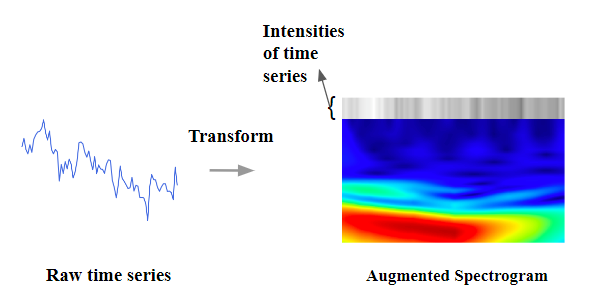}
    \caption{Visual representation of time series in the form of a time-frequency spectrogram augmented with intensities of time series at the top}
    \label{fig:teaser}
    \vspace{-0.5cm}
\end{figure}

In recent years, a new perspective has been introduced for forecasting time series where numeric data is converted to images to leverage successful computer vision algorithms for forecasting. These studies are motivated by the fact that traders' decision making is augmented by the visual representation of financial time series images such as charts, graphs, and plots. In addition, this approach can help to capture additional patterns and dependencies that are beneficial for understanding and forecasting time series data. However, these approaches typically use lineplots as the visual representation of time series, which misses some crucial information required for understanding time series. For example, frequency information is not apparent when directly looking at the time series in its raw form. 

One could use frequency spectrum of a time series as the visual representation. But this merely displays the various frequency components present in the data. High-frequency components are often associated with noise, while low-frequency components correspond to signals. However, it does not provide insights into the temporal dependencies between occurrences of different frequency components. The time-frequency spectrogram on the other hand visually represents how the frequency spectrum changes over time, enabling the learning of temporal dependencies between different frequency patterns.  


The main focus of this research is to use time-frequency spectrogram as visual representation of time series data and leveraging vision transformers to achieve simultaneous learning in both the time and frequency domains, enabling accurate forecasting. The paper presents the following key contributions:
\begin{itemize}
    \item We demonstrate the superiority of spectrograms as a visual representation of time series data, leading to enhanced predictions across various domains and diverse datasets.
    \item We leverage transformer's ability to learn cross-modality information across time and frequency domain to enhance the forecasting process.
\end{itemize}
Through comprehensive experiments conducted on diverse datasets from different domains, including synthetic, temperature, and financial time series, the paper demonstrates the advantages and effectiveness of the proposed approach in time series forecasting.
\section{Related Works}
\subsection{Time series forecasting}\label{ts_forecasting} 
Time series forecasting is a crucial task involving the statistical analysis of historical data to predict future values. In the literature, traditional forecasting techniques have relied on statistical tools like exponential smoothing~\cite{holt2004forecasting, winters1960forecasting, gardner1985forecasting} and autoregressive integrated moving average (ARIMA)~\cite{makridakis2020m4}, primarily applied to numerical time series data for one-step-ahead predictions. Moreover, several traditional multi-horizon forecasting methods~\cite{taieb2010multiple, marcellino2006comparison} have been developed to generate simultaneous predictions for multiple future time steps.

Machine learning (ML) approaches have shown promise in improving forecasting performance by effectively addressing high-dimensional and non-linear feature interactions in a model-free manner. These ML methods encompass tree-based algorithms, ensemble methods, neural networks, autoregression, and recurrent neural networks (RNNs)~\cite{hastie2001elements}. In recent works, deep learning (DL) methods have gained traction in time series forecasting, demonstrating impressive results on numerical time series data~\cite{bao2017deep, gensler2016deep, romeu2015stacked, sagheer2019time, Sutskever2014sequence, zhang2019deeplob}. DL techniques automate the feature extraction process, eliminating the need for domain expertise and offering promising outcomes in forecasting tasks.

In the domain of natural language processing (NLP), RNNs, long short-term memory (LSTM)~\cite{hochreiter1997long}, and gated recurrent units (GRUs)~\cite{cho2014learning} have been widely used for tasks like machine translation and language modeling. Interestingly, these models have been adapted for time series forecasting, where they have outperformed traditional statistical methods. Furthermore, the DeepAR algorithm, utilizing the RNN backbone, has emerged as a potent approach in time series forecasting. DeepAR has been more recently proposed and has shown superior performance compared to traditional forecasting techniques, particularly for datasets with multiple related time series~\cite{salinas2020deepar}. This highlights the potential of RNN-based methods in handling complex time series data and effectively capturing temporal dependencies.

Transformers~\cite{vaswani2017attention}, on the other hand, have revolutionized NLP applications and have also shown promise in time series forecasting. Unlike traditional RNN and LSTM networks, transformers leverage multi-headed self-attention to process all inputs simultaneously. This parallel computation capability not only reduces training time but also enables transformers to handle longer sequences without suffering from long-term memory dependency issues, often encountered in RNN-based models. These characteristics make transformers well-suited for handling time series forecasting tasks effectively~\cite{wu2020deep, zhou2021informer, zeng2023financial}. By effectively capturing complex temporal dependencies within time series data, transformers have demonstrated their ability to outperform traditional methods and are gaining popularity as a powerful tool in the field of time series forecasting. Given the promising merits of transformers in the context of time series forecasting, we will utilize transformers as the main model architecture in this work.

\subsection{Visual time series forecasting}
In recent years, a novel approach has gained traction in time series forecasting, which involves converting numeric data into images to leverage successful computer vision algorithms. This approach is motivated by the belief that visual representations of financial time series, such as charts and graphs, can enhance traders' decision-making. Prior research has explored representing time series data as images of line plots~\cite{sood2021visual, semenoglou2023image}, recurrent plots~\cite{li2020forecasting}, and candlestick plots~\cite{cohen2020trading} for both forecasting and classification tasks. Furthermore, recent advancements in video prediction from the computer vision domain have been adapted to improve forecasting accuracy~\cite{zeng2021deep}. These studies have bridged the gap between computer vision and time series forecasting, offering innovative ways to capture and analyze temporal patterns in numerical data.

Given the inherent unpredictability in financial time series, researchers have turned to decomposition techniques to reveal more detailed information at varying frequency levels~\cite{chaudhuri2016spectral} for better-informed portfolio management~\cite{scalzo2021nonstationary, arroyo2022dynamic} and risk modeling~\cite{BANDI2021214}. Recent works have explored time series forecasting or classification with frequency representation combined with deep learning~\cite{li2020forecasting, zhou2022fedformer, zhang2022self, moreno2023deep}. In our work, we used the wavelet transform as a time-frequency representation of the time series. The advantage of the wavelet transform over other frequency representations such as the Fourier transform lies in its ability to extract both local spectral and temporal information, resulting in a more comprehensive representation of the underlying data patterns. Researchers have already explored the utilization of time-frequency spectrograms for time series classification in the finance domain~\cite{du2020image}. In our study, we adopt the time-frequency spectrograms for a harder task on time series forecasting. Specifically, we leverage a multimodal image representation consisting of a time-frequency spectrogram augmented with intensities of numerical time series data, capitalizing on the advantages of this combined approach.

Moreover, transformers have been extended into the computer vision domain by integrating self-attention with convolutional neural network (CNN) architectures~\cite{dosovitskiy2020image}. Dosovitskiy et al. demonstrated that a pure transformer applied directly to sequences of image patches can achieve outstanding performance in image classification tasks. Recognizing this potential, our study adopts a vision transformer for simultaneous learning in both the time and frequency domains, capitalizing on the spectrogram representation of the time series data. The vision transformer's ability to handle cross-modality makes it well-suited for this task.

\section{Data}\label{sec:data}
This paper analyzes three datasets out of which one is synthetic (Section \ref{subsec:synthetic}) and the other two are real datasets (Section \ref{subsec:real temp} and \ref{subsec:real finance}).  This helps us to comprehensively evaluate spectrogram-based forecasting techniques as they encompass varying levels of periodicity and complexity, providing a comprehensive basis for the examination.

\subsection{Synthetic Data}\label{subsec:synthetic}
We created a synthetic dataset consisting of multiple periods by sampling data from harmonic functions. The initial dataset is artificially generated and aims to have complexity while still exhibiting a prominent and repeated signal. We synthesized the time series $s_t$ using a linearly additive two-timescale harmonic generating function:
\begin{equation}
s_t = (A_1 + B_1t)\sin(2\pi t/T_1 + \phi_1) + (A_2 + B_2t)\sin(2\pi t/T_2 + \phi_2)
\end{equation}
In the equation, the variable $t$ ranges from $t = 1$ to $t = T$, where $T$ represents the total length of the time series. The multiplicative amplitudes $A_1$ and $A_2$ are randomly selected from a Gaussian distribution $\mathcal{N}(1, 0.5)$. The amplitudes of the linear trends, $B_1$ and $B_2$, are sampled from a uniform distribution $\mathcal{U}(-1/T , 1/T)$. The driving time scales, $T_1$ and $T_2$, are relatively short and long compared to the total length $T$. Therefore, $T_1$ follows a normal distribution $\mathcal{N}(T/5, T/10)$, while $T_2$ follows a normal distribution $\mathcal{N}(T, T/2)$. Finally, the phase shifts $\phi_1$ and $\phi_2$ are chosen from a uniform distribution $\mathcal{U}(0, 2\pi)$.
Overall, we generated 150K time series of length 100 that were used for our experiments. Each time series differ concerning the possible combination of tuning parameters. The first panel in Figure \ref{fig:data_examples} displays a sampled instance of the synthetic data. It is evident that the synthetic time series comprises two distinct time scales: brief oscillations superimposed on much longer wave trains.

\subsection{Temperature data}\label{subsec:real temp}
This dataset contains various temperature observations gathered by the Australian Bureau of Meteorology for 422 weather stations across Australia, between 02/05/2015 and 26/04/2017. The dataset was obtained from Monash repo~\cite{godahewa2021monash}. The dataset contains several attributes as equal lenght time series (725 days).  For our experiments, we used the daily mean temperature in Celsius for each station recorded over a 24-hour period.
For every station, we sampled 30 time series of length 60 for every station.  Overall We had 7320 time series. The second panel in Figure ~\ref{fig:data_examples} displays a sampled instance of the temperature data.

\subsection{Financial data}\label{subsec:real finance}
We obtained the daily stock prices of S\&P 500 constituent stocks through the Yahoo! finance database. The dataset comprises daily Adjusted Close values for the stocks that constitute the S\&P-500 index, spanning from the year 2000 onwards. In total, our dataset includes around 58K individual time series, with each time series segment consisting of 100 days of data. The third panel in Figure ~\ref{fig:data_examples} displays a
sampled instance of the financial data.

\section{Method}\label{sec:method}
\begin{figure*}[t!]
    \includegraphics[scale = 0.48]{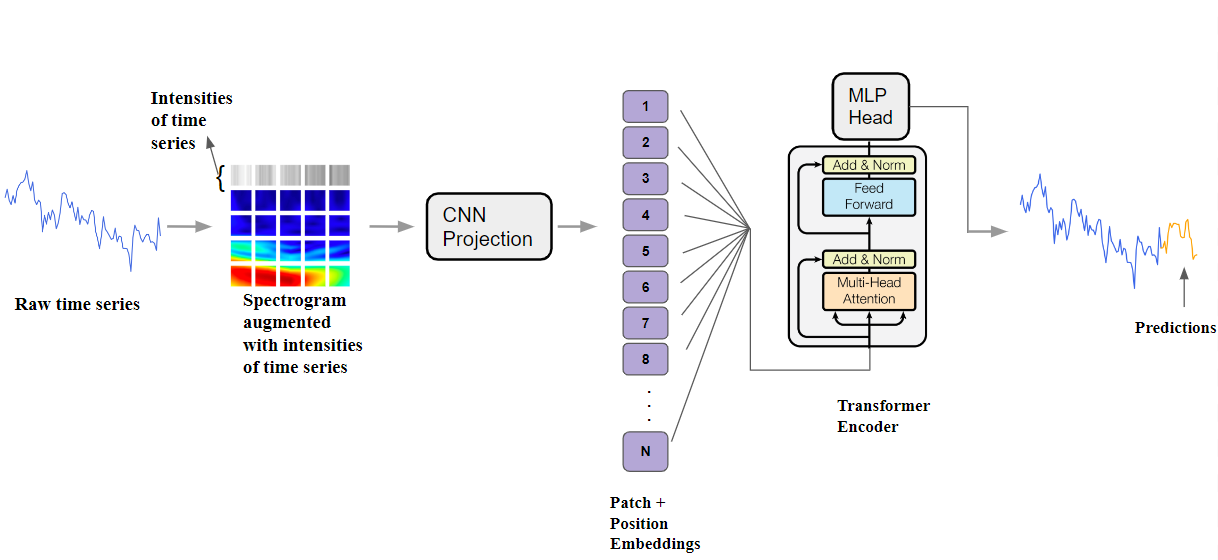}
    \caption{Overview of the proposed approach.}
    \label{fig:overview}
\end{figure*}

Our goal is time series forecasting using a \textbf{\underline{vi}}sion \textbf{\underline{t}}ransformer and a multimodal image of time-frequency \textbf{\underline{spec}}trogram augmented with intensities of \textbf{\underline{num}}erical time series, which we refer to as \textbf{ViT-num-spec}. This approach leverages the strengths of both the visual representation from the spectrogram and the numerical information to improve forecasting performance. The approach involves converting the numeric time series into an image using a time-frequency spectrogram (Section \ref{subsec:spectrogram}), and then employing a vision transformer encoder appended with a multi-layer perceptron (MLP) head to forecast the future (Section \ref{subsec:forecast_vision_transformer}), as illustrated in Figure~\ref{fig:overview}.

\subsection{Preprocessing}\label{subsec:preprocess}
The temperature dataset contained missing values, which were handled by performing forward filling. This means that any missing values were replaced with the most recent available value in the dataset, moving forward in time.
For each of the three datasets, the data $\textbf{x}$ was scaled to ensure that the scaled $\textbf{x}$ fell within the range of $[0, 1]$. This scaled $\textbf{x}$ was then utilized to incorporate the original time series information into the spectrogram and to facilitate the learning of the target, as explained in more detail later.

\begin{figure}[t!]
    \centering
    \includegraphics[scale = 0.28]{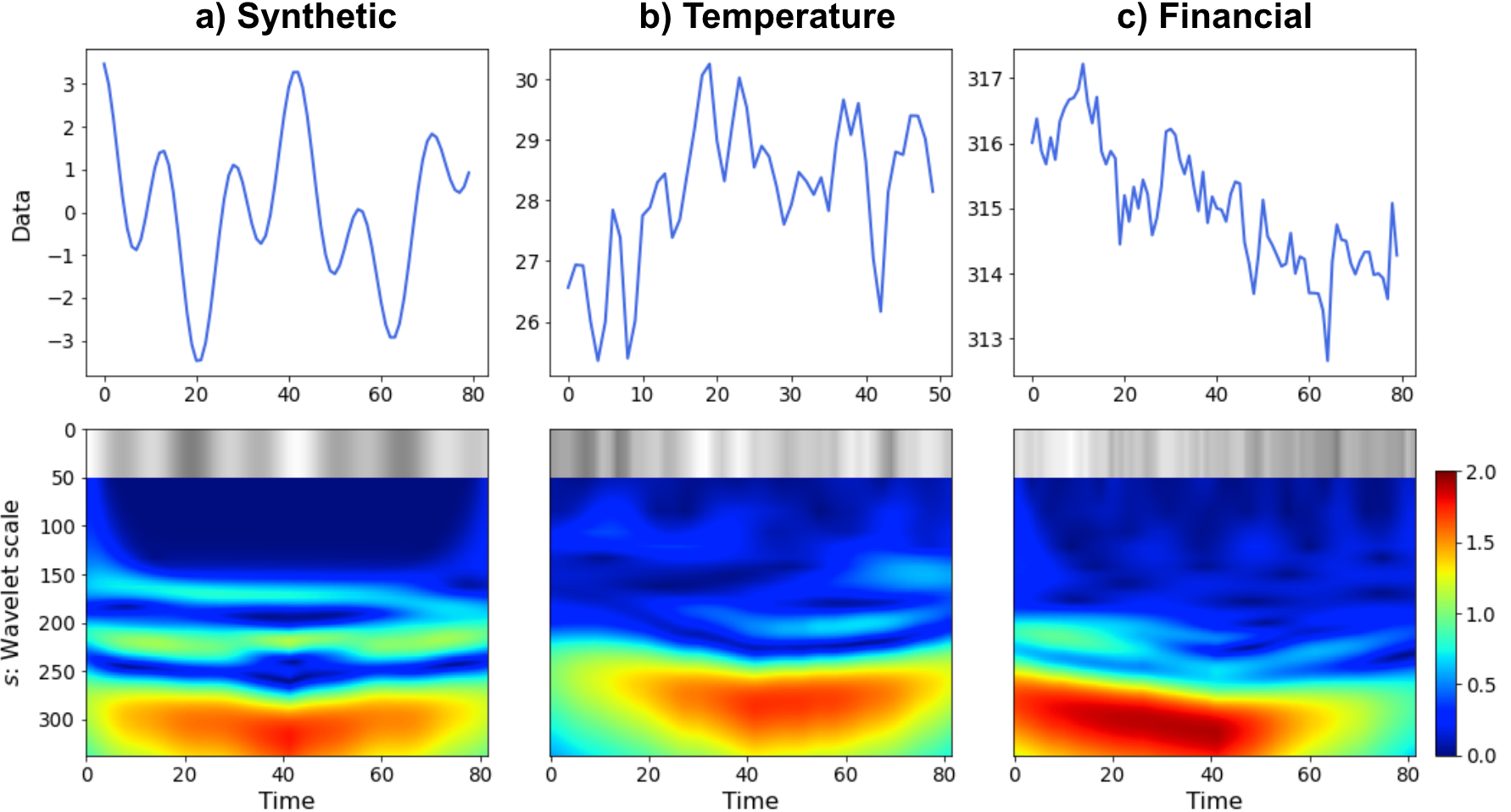}
    \caption{Illustrations of the inputs of the three datasets: a) synthetic, b) temperature, and c) financial stock prices. The top panels show the raw time series represented as lineplots and the bottom panels depict the augmented time-frequency spectrogram. Each input time series consists of 80 steps for the synthetic and financial datasets, while the temperature dataset has 50 steps. For the financial and temperature data, each time step represents a 1-day time interval. }
    \label{fig:data_examples}
     \vspace{-0.5cm}
\end{figure}

\subsection{Time-Frequency Spectrogram} \label{subsec:spectrogram}
To generate a time-frequency spectrogram from a given time series, two common methods are Short-Time Fourier transforms (STFT)~\cite{griffin1984signal} and wavelet transforms~\cite{daubechies1990wavelet}. STFT involves a sliding window-based version of the Fourier transform, extracting frequency components within a fixed time window. However, it can only uncover a limited set of frequency components within this fixed window and may not be suitable for non-periodic and transient signals in stock price time series. On the other hand, wavelet transform is well-suited for analyzing time series with transient signals, as it can uncover varying frequency components~\cite{frisch1992use, raihan2005wavelet}. Unlike STFT, wavelet transform is not limited by a fixed time window, making it a more suitable choice for our study. Therefore, we choose to use wavelet transforms in our approach for generating time-frequency spectrograms from time series data.

A wavelet is a wave-like oscillation (zero-mean signal) that is localized in both time and frequency space. For our approach, we utilized the morlet wavelet (see Equation \eqref{wavelet_eq}), which has been demonstrated to be effective in time series classification~\cite{du2020image}.
\begin{equation}\label{wavelet_eq}
\psi(x) = \sqrt{\frac{1}{s}} \pi^{-\frac{1}{4}} \mathrm{e}^{\frac{1}{2}\frac{x}{s}} \mathrm{e}^{jw\frac{x}{s}},
\end{equation}
where $s$ represents the scale of the wavelet. The morlet wavelet is essentially a sine wave multiplied by a Gaussian envelope centered at zero. The scale $s$ determines the width and frequency of the wavelet. As $s$ increases, the frequency of the wavelet becomes wider and lower.
The wavelet transform is performed by convoluting wavelets at different scales with the time series. The magnitude of the resulting multiplication coefficients represents the signal strength of different wavelets. This is saved as a heatmap (or spectrogram), as depicted in Figure~\ref{fig:data_examples}. The rows of the spectrogram correspond to varying wavelet scales, i.e., varying frequencies. The columns correspond to time. Higher frequency components are at the top, and lower frequency components are at the bottom as the $s$ increases from the top to the bottom.

Note that we augmented the spectrogram by adding an image stripe row at the top, and the resulting full image was used as the input image to our model. Throughout the paper, we will continue to refer to the augmented image as the spectrogram. The reason for this augmentation is as follows: the standard spectrogram only visualizes the strength of varying signals, but it does not retain the sign of the signal, which is available in the time domain. To preserve the sign information, we convert the standardized time series to an image row with intensities represented as integers in the range of $[0, 255]$. Examples of time-frequency spectrograms are provided in Figure~\ref{fig:data_examples}.

\subsection{Vision Transformer for Forecasting}\label{subsec:forecast_vision_transformer}
We utilize a vision transformer encoder with an appended MLP head for time series forecasting, as illustrated in Figure~\ref{fig:overview}. To effectively process the input image, we divide it into non-overlapping patches of equal size. This division results in the horizontal time axis being split into image-patch-sized time intervals. These patches are then converted to tokens through linear projection, and standard 1D position embedding is added. The objective of the forecasting model is to learn temporal dependencies between time and frequency patterns across the horizontal time axis. The encoder learns to encode these patches into latent feature vectors. In our experiments, we use an input image size of $128 \times 128$, and image patches of size $16 \times 16$. The top row of the price time series is scaled to $16 \times 128$, and the spectrogram is scaled to $112 \times 128$, ensuring that they are divisible by the patch size. The forecast network is constructed by appending a lightweight MLP head to the vision transformer encoder, as depicted in Figure~\ref{fig:overview}. Further details on the network architecture and training can be found in Section \ref{experiments_models}.

\section{Experiments}\label{sec:experiments}
In our experimental evaluation, we compared the performance of seven methods: 1) ViT-num-spec - the proposed method utilizing the vision transformer and a multimodal image of time-frequency spectrogram augmented with intensities of numerical time series, 2) ViT-lineplot - utilizing the vision transformer on a lineplot visualization of the time series, 3) ViT-num - an ablation study on using only the intensities of time series as input, 4) DeepAR - a state-of-the-art LSTM-based time series forecasting method, as well as statistical baselines such as 5) EMA, and 6) ARIMA and 7) naive numeric forecasting baseline. The evaluation was conducted using two widely used metrics, namely SMAPE and MASE, to assess performance.

Along with traditional numerical time series forecasting methods, we showcased the advantages of using a visual spectrogram as a representation for time series data compared to other visual representations and an ablation study using only the intensities of time series as input. Additionally, we highlighted the benefits of employing a vision transformer for simultaneous learning in both the time and frequency domains. The experiments were conducted on a Linux machine equipped with 8 NVIDIA T4 Tensor Core GPUs, each with 16GB memory. PyTorch v1.0.0 DL platform in Python 3.6 was used for all models. To ensure reproducibility, we set a fixed random seed of 42.

\subsection{Experimental setup}\label{subsec:setup}
\subsubsection{Synthetic Data}
Our training dataset comprised 80K samples, while the validation dataset consisted of 20K samples. For the testing phase, we had a dataset of 50K samples available. Each time series consists of 100 time steps, and the task involved predicting the subsequent 20 time steps using the initial 80 time steps as input for each time series.

\subsubsection{Temperature Data}
For training purposes, data from 2015 and 2016 was utilized, while data from 2017 was reserved for testing. The construction of the training, testing, and validation datasets involved randomly sampling 10 time series of lenght 60 per station. Therefore, each of the train, test, and val datasets consisted of 4220 time series of length 60. The forecasting task entailed using the initial 50 time steps as the input to predict the subsequent 10 time steps for each individual time series. We maintained the length of the past timeseries as 50 to forecast the next 10 steps due to the small size of the daily temperature dataset, and to minimize the overlap in the generated time series. This means that we utilized the temperature data from the past 50 days to predict the temperature for the upcoming 10 days.

\subsubsection{Financial Data} For the training data, we randomly sampled information from the period between 2000 and 2014. As for the test data, we sampled information from the years 2016 to 2019. The training set consists of a total of 46,875 time series with a 0.8 train and validation split, and we sampled 15,625 time series as the test set. The forecasting task involved using the first 80 time steps to predict the subsequent 20 time steps for each time series. 

For each of the three datasets, the performance evaluation involved averaging metrics over the 20 predicted time steps for the synthetic and financial datasets, and over 10 predicted time steps for the temperature dataset. The overall performance on the test set is represented by the mean and standard deviation of the evaluation metrics calculated across the entire testing set for each dataset.

\subsection{Models}\label{experiments_models}
\subsubsection{ViT-num-spec}\label{subsubsec:vit}
ViT-num-spec is our proposed method that utilizes the vision transformer on a multimodal image of time-frequency spectrogram augmented with intensities of numerical time series as the model input. The latent embedding vector has a dimension of 128. We trained the model using a batch size of 128 and the AdamW optimizer with a weight decay of 0.05. The training process was performed for a maximum of 200 epochs, and we implemented an early stopping mechanism with a patience of 10. The base learning rate and learning rate scheduler hyperparameters were tuned for each dataset.

\subsubsection{ViT-lineplot} \label{subsubsec:vit_lineplot}
ViT-lineplot is a variant of our proposed method that utilizes the vision transformer on a lineplot visualization of the time series, without including the frequency information. It shares the same architecture with ViT-num-spec, including the loss function. Similar to ViT-num-spec, the base learning rate and learning rate scheduler hyperparameters were tuned for each dataset in the case of ViT-lineplot as well.

\subsubsection{ViT-num}\label{subsubsec:vit_num}
ViT-num is an ablation study of our proposed method that uses only the intensities of time series as input, specifically the top price row of a spectrogram image. Similar to ViT-lineplot, ViT-num shares the same architecture with ViT-num-spec, and the learning rate and scheduler were tuned for this variant as well.


\subsubsection{DeepAR} \label{sec:deepar}
DeepAR is an effective supervised learning algorithm used for forecasting univariate time series \cite{salinas2020deepar}.  It leverages autoregressive recurrent neural networks, specifically built on LSTM architecture, to generate predictions. DeepAR generates probabilistic forecasts by training a single model simultaneously on a vast array of interconnected time series. In our implementation, we utilized DeepAR to forecast the value distribution for the next 20 steps in the harmonic and financial datasets, while using an input time series length of 80. For the temperature dataset, we used an input length of 50 to predict the next 10 steps. We utilized the normal distribution loss and employed the AdamW optimizer with a batch size of 128. The model underwent training for a maximum of 300 epochs, incorporating an early stopping mechanism with a patience of 15. The initial learning rate was set at 1e-3 and adjusted by a learning rate scheduler with a decay factor of 0.1 and a patience of 5. To prevent overfitting, a drop rate of 0.1 was applied. DeepAR generated 200 samples of the prediction target for each time series, and the final prediction was derived by calculating the average across these predicted samples.

\subsubsection{ARIMA} \label{sec:arima}
ARIMA (AutoRegressive Integrated Moving Average) is a forecasting model that combines autoregressive (AR), differencing (I), and moving average (MA) components to capture dependencies and patterns in time series data, enabling predictions based on past observations \cite{wilks2011statistical}. By incorporating lagged values and previous errors, ARIMA models effectively capture trends, seasonality, and other temporal patterns.

To streamline the modeling process, we utilized the auto-ARIMA procedure \cite{autoarima}. This automated approach employs algorithms to systematically search for the optimal ARIMA parameter configuration, utilizing the Akaike Information Criterion. This automated parameter selection saves time and eliminates the need for manual tuning. Additionally, we implemented the stepwise algorithm described in \cite{hyndman2008automatic} to further enhance the efficiency of the model selection process.
\subsubsection{EMA} \label{sec:ema}
Exponential Moving Average (EMA) is a smoothing technique commonly employed in data analysis and forecasting. EMA distinguishes itself from the simple moving average by giving more significance to recent observations through the use of varying weights assigned to data points. This weighting scheme grants EMA the ability to swiftly detect and highlight short-term trends in the data. By iteratively updating the average, EMA combines the previous EMA value with the current observation, utilizing a smoothing factor known as the smoothing constant. This constant provides control over the level of responsiveness and smoothness exhibited by the EMA.
\subsubsection{Naive} \label{sec:naive}
We developed a simplistic baseline model that relied on the assumption that the last observed value would serve as the prediction for all future time steps.

\begin{table*}[t!] 
\centering
\resizebox{\textwidth}{!}{%
\begin{tabular}{>{\centering\arraybackslash} c| >{\centering\arraybackslash}c >{\centering\arraybackslash}c >{\centering\arraybackslash}c>{\centering\arraybackslash}c|>{\centering\arraybackslash}c>{\centering\arraybackslash}c>{\centering\arraybackslash}c } 
\hline
& \multicolumn{4}{c|}{\textbf{Baselines}} & \multicolumn{3}{c}{\textbf{ViT}}  \\ \hline 
\textbf{SMAPE $\downarrow$} & Naive & EMA & ARIMA & DeepAR & ViT-num & ViT-lineplot & \textbf{ViT-num-spec
}
\begin{tabular}[c]{@{}c@{}}\end{tabular} \\ \hline
Synthetic & 1.234 $\pm$ 0.442 & 1.242 $\pm$ 0.447 & 0.549 $\pm$ 0.379 & 1.073 $\pm$ 0.403 & 0.376 $\pm$ 0.294 & 0.531 $\pm$ 0.346 & \textbf{0.304 $\pm$ 0.258} \\ 
Temperature & 0.134 $\pm$ 0.110 & 0.132 $\pm$ 0.109 & 0.136 $\pm$ 0.109 & 0.128 $\pm$ 0.102 & 0.132 $\pm$ 0.104 & 0.135 $\pm$ 0.107 & \textbf{0.125 $\pm$ 0.100} \\  
Financial & 0.036 $\pm$ 0.028 & 0.036  $\pm$ 0.028 & 0.041 $\pm$ 0.035 & 0.038 $\pm$ 0.030 & 0.036 $\pm$ 0.028 & 0.068 $\pm$ 0.054 & 0.036 $\pm$ 0.028 \\ 

\hline
\textbf{MASE $\downarrow$} &  &  &  &  &  &  & 
\begin{tabular}[c]{@{}c@{}}\end{tabular}
\\ 
\hline
Synthetic & 5.272 $\pm$ 3.892 & 5.265 $\pm$ 3.899 & 2.194 $\pm$ 2.679 & 4.225 $\pm$ 4.092 & 0.921 $\pm$ 1.078 & 1.518 $\pm$ 1.704 & \textbf{0.757 $\pm$ 0.974}\\
Temperature & 1.497 $\pm$ 0.885 & 1.460 $\pm$ 0.862 & 1.499 $\pm$ 0.818 & 1.446 $\pm$ 0.792 & 1.513 $\pm$ 0.843 & 1.526 $\pm$ 0.904 & \textbf{1.413 $\pm$ 0.793}\\ 
Financial &  3.486 $\pm$ 2.789 & 3.487 $\pm$ 2.788 & 3.884 $\pm$ 3.076 & 3.659 $\pm$ 2.905 & 3.464 $\pm$ 2.796 & 6.383 $\pm$ 4.344 & \textbf{3.461 $\pm$ 2.801}\\ \hline
\textbf{Sign accuracy $\uparrow$} &  &  &  &  &  &  & 
\begin{tabular}[c]{@{}c@{}}\end{tabular}
\\ 
\hline
Synthetic &  0.0\% & 44.3\% & 88.0\% & 75.1\% & 94.4\% & 90.9\% & \textbf{95.7\%}\\
Temperature &  0.2\% & 59.5\% & 60.4\% & \textbf{65.1\%} & 61.3\% & 60.7\% & 64.2\%\\
Financial &  0.3\% / 0.0\%
 & 49.7\% / 50.7\% & 50.6\% / 50.9 & 51.5\% / 53.0\%
 & 54.3\% / 57.4\% & 50.3\% / 50.5\%
 & \textbf{54.4\% / 58.4\%}\\ \hline
\end{tabular}
}
\caption{Summary of the evaluation metrics:
For the financial dataset, we report both the original/thresholded sign accuracies (threshold is 20\% of the standard deviation of past time series data). The results are reported as the mean $\pm$ standard deviation of the evaluation metrics calculated across the entire testing set. }
\label{tab:results}
 \vspace{-0.5cm}
\end{table*}

\subsection{Evaluation metrics}
To evaluate the performance of our models, the forecasted prices were initially transformed back to the original scale. All of our models produced continuous predictions, and we assessed their performance using three metrics: SMAPE (Symmetric Mean Absolute Percentage Error),  MASE (Mean Absolute Scaled Error) and sign accuracy.
\begin{itemize}
    \item[1)] \textbf{SMAPE:} The symmetric mean absolute percentage error (SMAPE) is defined as, 
    \begin{equation} \label{eq:smape}
        SMAPE = \frac{1}{T}\sum_{t=1}^T\frac{|x_t - \hat{x}_t|}{(|x_t| +|\hat{x}_t|)/2}, 
    \end{equation}
where $T$ denotes the length of the predicted time series (in our case, $T$ = 1, 2, 3, 4, 5); $x_t$ and $\hat{x}_t$ is the observed ground truth and it's corresponding forecast at time step $t$. SMAPE values lie within [0, 2], with smaller values imply more accurate forecast.  

\item [2)] \textbf{MASE:} The mean absolute scaled error (MASE) is defined as,
\begin{equation}\label{eq:mase}
    MASE=\frac{\frac{1}{T} \sum_{i=t+1}^{t+T}|x_{i} - \hat{x_{i}}|}{\frac{1}{t}\sum_{i=1}^{t}|x_{i} - x_{i-1}|}
\end{equation}
It is the mean absolute error of the forecast divided by the mean absolute error of naive one-step forecast on the in-sample data. Note that MASE becomes unstable when the denominator approaches 0. Similar to SMAPE, lower MASE values indicate better forecasts.
\item[3)] \textbf{Sign Accuracy:} This represents the average classification accuracy across all test samples. A higher accuracy indicates more precise forecasts. We define three classes to classify forecast movements: class 1 for "going up", class 2 for "remaining flat," and class 3 for "going down". To determine the class, we compare the difference between the predicted forecasts and the last observed value in the input time series.

\end{itemize}


\subsection{Results}
\begin{figure*}[htb]
    \centering
    \includegraphics[scale = 0.5]{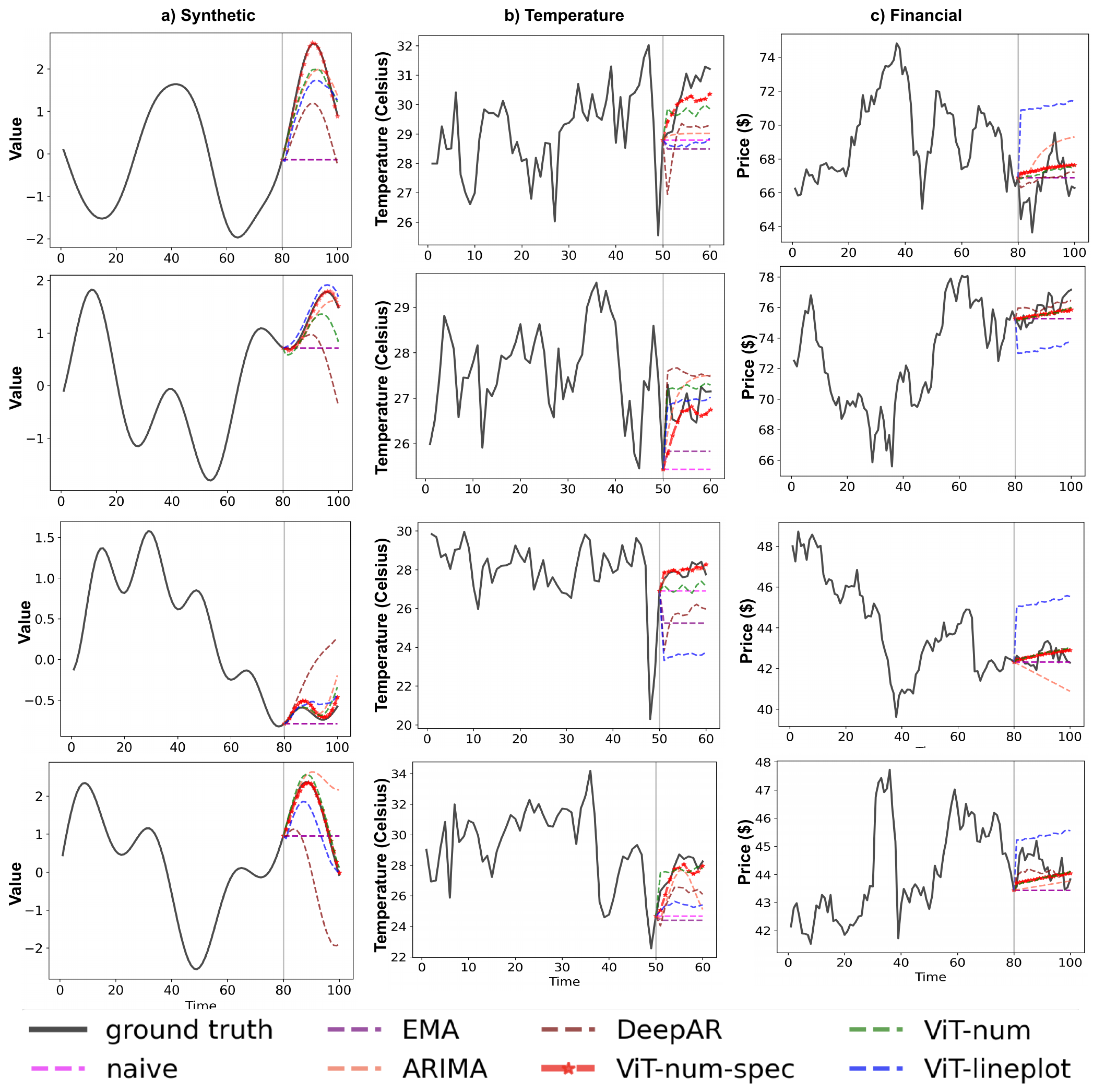}
    \caption{Qualitative examples for predictions for the three datasets: a) synthetic, b) temperature, and c) financial stock prices.
    }
    \label{fig:qualitative_results}
     \vspace{-0.5cm}
\end{figure*}

We present a summary of our quantitative benchmark results in Table~\ref{tab:results}, which includes the mean and standard deviation of the evaluation metrics calculated across the entire testing set. Additionally, we provide qualitative examples comparing different methods across all three datasets in Figure~\ref{fig:qualitative_results}.


Overall, our proposed approach, ViT-num-spec, demonstrates its superiority over all state-of-the-art statistical and deep learning baselines, firmly establishing its competitiveness in the field of numerical time series forecasting. Notably, it outperforms ViT-lineplot and ViT-num, thus highlighting the advantages of utilizing a multimodal image of time-frequency spectrogram augmented with intensities of numerical time series as the visual representation for time series data.
The successful fusion of the multimodal image representation and the vision transformer's simultaneous learning in both the time and frequency domains plays a crucial role in significantly enhancing the forecasting performance. This distinct capability sets ViT-num-spec apart from traditional time series forecasting methods, providing it with a clear advantage in effectively handling complex temporal dependencies. In the following sections, we provide a dataset-wise breakdown of the results to offer a comprehensive understanding of ViT-num-spec's performance across different datasets.

\subsubsection{Synthetic data}

Among all the evaluated methods, all the ViT variants outperform all baseline models.  This shows that a vision transformer model can successfully capture the patterns present in the input signal. Among the Vit methods, ViT-num-spec performing the best, followed by ViT-num, and then ViT-lineplot. Moreover, ViT-num-spec's superiority over ViT-num demonstrates the benefits of using a multimodal setup that includes both spectrogram and intensities. Lastly, ViT-num-spec's outperformance over ViT-lineplot underscores the superiority of using time-frequency spectrograms instead of simple lineplot visualizations for representing time series data.
All these findings emphasizes the advantage of utilizing multimodal representations for time series forecasting, where the combination of spectrogram and intensities contributes to improved performance in capturing complex temporal patterns.

\subsubsection{Temperature data}

Among all the evaluated methods, ViT-num-spec emerges as the top performer in terms of SMAPE and MASE, 
However, DeepAR achieves the highest accuracy in sign predictions. Note that the sign accuracy of ViT-num-spec is better than other ViT variants. While sign accuracy is crucial for making correct directional predictions, SMAPE and MASE are considered more crucial for achieving precise temperature forecasts. The performance of ViT-num-spec is still better than other ViT variants.
This trend is consistent with what was observed in the synthetic datasets. This further highlights the significance of incorporating a vision transformer in a multimodal setup that includes both spectrogram and intensities.


\subsubsection{Financial data}
Among the three datasets, financial time series present the most significant forecasting challenges. Prior research~\cite{pedersen2019efficiently} indicates that financial data exhibits properties resembling randomness at short scales and lacks evident periodic patterns at larger scales. When considering SMAPE and MASE, all methods demonstrate similar performance, except for ViT-lineplot, which performs the worst among the evaluated methods. Upon analyzing sign accuracy, it becomes evident that ViT-num-spec and ViT-num outperform the other methods.  Thus, lineplots are not a good visual representation of financial data. For practical decision-making in financial scenarios, buying or selling actions are usually executed only when the predicted change exceeds a certain significant threshold. To address this, we apply a thresholding technique where the threshold value is set to 20\% of the standard deviation of the past time series data. By using this threshold, we evaluate the sign accuracy of the predictions, which provides a more realistic representation of the model's performance in making actionable decisions for financial forecasting.  Indeed, when considering the threshold accuracy, ViT-num-spec emerges as the top performer, closely followed by ViT-num. This reinforces the advantage of using spectrograms over lineplots for representing time series data, as demonstrated by the superiority of both ViT-num-spec and ViT-num over ViT-lineplot. Furthermore, the ground truth statistics reveal that the distribution of positive, negative, and neutral signs in the dataset is 55.7\%, 44.0\%, and 0.3\% respectively. When considering the dominant trend as the forecasted sign, we achieve an accuracy of 55.7\%, which is lower than ViT-num-spec's performance after applying thresholding.

In summary, the results validate the effectiveness of our proposed approach, which combines the multimodal image representation and the vision transformer's simultaneous learning in both the time and frequency domains. This fusion plays a crucial role in significantly enhancing the forecasting performance, demonstrating the capability of achieving state-of-the-art results in time series forecasting.


\section{Conclusion}\label{sec:conclusion}
In conclusion, this paper introduces a novel approach for time series forecasting, utilizing time-frequency spectrograms as visual representation of raw time sereis and incorporating a vision transformer for multimodal learning. We compare our method against benchmarked methods, including statistical baselines and state-of-the-art deep learning-based approaches.  We also conduct experiments using lineplot as the visual representation of time series and an ablation study using only the intensities of time series 
 to demonstrate the superiority of our proposed approach in forecasting across diverse datasets, including synthetic, temperature, and financial stock price data.


\begin{acks}
This paper was prepared for informational purposes by the Artificial Intelligence Research group of JPMorgan Chase \& Co and its affiliates (“J.P. Morgan”) and is not a product of the Research Department of J.P. Morgan.  J.P. Morgan makes no representation and warranty whatsoever and disclaims all liability, for the completeness, accuracy or reliability of the information contained herein.  This document is not intended as investment research or investment advice, or a recommendation, offer or solicitation for the purchase or sale of any security, financial instrument, financial product or service, or to be used in any way for evaluating the merits of participating in any transaction, and shall not constitute a solicitation under any jurisdiction or to any person, if such solicitation under such jurisdiction or to such person would be unlawful. 
\end{acks}


\bibliographystyle{ACM-Reference-Format}
\bibliography{egbib}

\end{document}